\documentclass[sigconf]{acmart}

\AtBeginDocument{%
  \providecommand\BibTeX{{%
    \normalfont B\kern-0.5em{\scshape i\kern-0.25em b}\kern-0.8em\TeX}}}

\copyrightyear{2022} 
\acmYear{2022} 
\setcopyright{acmlicensed}\acmConference[GECCO '22]{Genetic and Evolutionary Computation Conference}{July 9--13, 2022}{Boston, MA, USA}
\acmBooktitle{Genetic and Evolutionary Computation Conference (GECCO '22), July 9--13, 2022, Boston, MA, USA}
\acmPrice{15.00}
\acmDOI{10.1145/3512290.3528690}
\acmISBN{978-1-4503-9237-2/22/07}

\usepackage{url}            
\usepackage{booktabs}       
\usepackage{amsfonts}       
\usepackage{nicefrac}       
\usepackage{microtype}      
\usepackage{xcolor}         
\usepackage{algorithm}
\usepackage{algorithmic}
\usepackage{amsmath}
\usepackage{mathtools}
\usepackage{bm}
\usepackage{subcaption}

\providecommand{\argmin}{\operatornamewithlimits{argmin}} 
\DeclareMathOperator{\Tr}{Tr}     
\DeclareMathOperator{\diag}{diag} 
\DeclareMathOperator{\vect}{vec}  
\providecommand{\R}{\mathbb{R}} 
\providecommand{\E}{\mathbb{E}} 
\providecommand{\T}{\mathrm{T}} 
\providecommand{\ind}[1]{\mathbb{I}\{#1\}} 
\renewcommand{\geq}{\geqslant} 
\renewcommand{\leq}{\leqslant} 
\DeclarePairedDelimiterX{\inner}[2]{\langle}{\rangle}{#1, #2}
\DeclarePairedDelimiter{\norm}{\lVert}{\rVert}
\DeclarePairedDelimiter{\abs}{\lvert}{\rvert}

\usepackage{amsthm}

\usepackage{cleveref}

\newtheorem{theorem}{Theorem}[section]
\newtheorem{proposition}[theorem]{Proposition}

\newtheorem{lemma}[theorem]{Lemma}
\theoremstyle{definition}

\newtheorem{assumption}[theorem]{Assumption}

\begin{document}

\title{Monotone Improvement of Information-Geometric Optimization Algorithms with a Surrogate Function}

\author{Youhei Akimoto}
\email{akimoto@cs.tsukuba.ac.jp}
\orcid{0000-0003-2760-8123}
\affiliation{%
  \institution{University of Tsukuba \& RIKEN AIP}
  \streetaddress{1-1-1 Tennodai}
  \city{Tsukuba}
  \state{Ibaraki}
  \country{Japan}
  \postcode{}
}


\begin{abstract}
A surrogate function is often employed to reduce the number of objective function evaluations for optimization. However, the effect of using a surrogate model in evolutionary approaches has not been theoretically investigated. This paper theoretically analyzes the information-geometric optimization framework using a surrogate function. The value of the expected objective function under the candidate sampling distribution is used as the measure of progress of the algorithm. We assume that the surrogate function is maintained so that the population version of the Kendall's rank correlation coefficient between the surrogate function and the objective function under the candidate sampling distribution is greater than or equal to a predefined threshold. We prove that information-geometric optimization using such a surrogate function leads to a monotonic decrease in the expected objective function value if the threshold is sufficiently close to one. The acceptable threshold value is analyzed for the case of the information-geometric optimization instantiated with Gaussian distributions, i.e., the rank-$\mu$ update CMA-ES, on a convex quadratic objective function. As an alternative to the Kendall's rank correlation coefficient, we investigate the use of the Pearson correlation coefficient between the weights assigned to candidate solutions based on the objective function and the surrogate function. 
\end{abstract}



\maketitle
\sloppy
\section{Introduction}

The covariance matrix adaptation evolution strategy (CMA-ES) \cite{rankonecma,rankmucma,cmamultimodal,ddcma,activecma} is a state-of-the-art approach for the minimization of a derivative-free black-box objective function $f:\R^d \to \R$ on a continuous domain. 
At each iteration, the CMA-ES samples multiple candidate solutions from a Gaussian distribution. These solutions are then evaluated on the objective function, and their rankings are computed. 
The distribution parameter of the Gaussian distribution is updated using the candidate solutions and their rankings.

When the objective function is computationally expensive to evaluate, a surrogate function $g:\R^d \to \R$ that is relatively computationally inexpensive to evaluate is often employed to reduce the number of $f(x)$ evaluations for optimization \cite{selfsurrogate,lmm-well,surveyexpensive}.
In the CMA-ES, the surrogate function $g$ is employed to approximate the rankings of the candidate solutions.
If the surrogate function $g$ provides an adequate approximation or the ground truth of the ranking of the candidate solutions as those based on $f$, it is expected that the behavior (i.e., the update of the distribution parameter) of the CMA-ES using this surrogate function is close to or identical to that of the CMA-ES using the ground truth objective function $f$, respectively. However, using the surrogate function can decrease the execution time.

Kendall's rank correlation coefficient $\tau$ \cite{kendall} is often employed to measure the quality of a surrogate function $g$ as an approximation of $f$ for use in the CMA-ES with a surrogate function \cite{lqcmaes,multifidelity,minmax}.
The $\tau$ between the objective function values $\{f(x_i)\}$ and the surrogate function values $\{g(x_i)\}$ of (or a subset of) candidate solutions are typically computed. 
If $\tau \geq \bar\tau$, for a pre-defined threshold $\bar\tau \in [-1, 1]$, the surrogate $g$ is used. 
If $\tau < \bar\tau$, the surrogate function $g$ is trained (in the case of surrogate-assisted optimization \cite{lqcmaes}) or refined (in the case of multi-fidelity optimization \cite{multifidelity} or min--max optimization \cite{minmax}) so that $\tau \geq \bar{\tau}$ is satisfied.
This forces the surrogate function $g$ to provide a good approximation of the rankings of the candidate solutions. 

The hypothesis behind this use of a surrogate function $g$ with Kendall's rank correlation coefficient $\tau$ is that \emph{similar rankings of candidate solutions result in similar updates of the distribution parameters}, where the similarity is measured by $\tau$. This hypothesis is supported indirectly and empirically by the successful applications of such surrogate-assisted approaches \cite{lqcmaes,multifidelity,constrainedmultifidelity,minmax,evolutioncontrol}. 
However, the effect of the use of a surrogate model in evolutionary approaches has been theoretically less investigated in the literature, as pointed out in \cite{surrogatereview}.
Probably, the most relevant existing studies are \cite{surrogatees1,surrogatees2}. 
In \cite{surrogatees1,surrogatees2}, the (1+1)-ES with a surrogate was analyzed on the spherical objective function $f(x) = \frac12 \norm{x}^2$, where the surrogate function is assumed to follow the normal distribution with center $f(x)$ and a fixed standard deviation. 
The quality of the surrogate function was measured by the standard deviation employed in \cite{surrogatees1,surrogatees2}; in contrast, herein, we are more interested in the effect of the rank correlation on the current work as it is used in practice.

In this paper, we theoretically investigate the effect of a surrogate function in the framework of information-geometric optimization (IGO) \cite{igo}. 
IGO is a generic mathematical framework of stochastic and comparison-based approaches, including a variant of the CMA-ES, namely the rank-$\mu$ update CMA-ES. 
IGO evolves a probability distribution $P_\theta$ with distribution parameter $\theta \in \Theta$ from which candidate solutions are sampled.
We use the expected objective function value $J(\theta) := \E_{x \sim P_{\theta}}[f(x)]$ to measure the progress of the IGO update. 
The main research question of this study is stated as follows:
Provided that $\E[J(\theta^{(t+1)}) \mid \theta^{(t)}] < J(\theta^{(t)})$ if the IGO update is performed using the objective function $f$, how large does Kendall's $\tau$ need to be to guarantee $\E[J(\theta^{(t+1)}) \mid \theta^{(t)}] < J(\theta^{(t)})$ if the IGO update is performed using surrogate function $g$?

The contributions of this paper are summarized as follows:
\begin{itemize}
\item We derive a sufficient condition to guarantee $\E[J(\theta^{(t+1)}) \mid \theta^{(t)}] < J(\theta^{(t)})$ when the IGO update is performed using surrogate function $g$ (\Cref{prop:fg}).
    
\item We show that if $\E[J(\theta^{(t+1)}) \mid \theta^{(t)}] < J(\theta^{(t)})$ is satisfied by the IGO update using the ground truth objective function $f$, then the IGO update using the surrogate function $g$, such that the population Kendall's $\tau$ between $f$ and $g$ is sufficiently high, satisfies the sufficient condition for $\E[J(\theta^{(t+1)}) \mid \theta^{(t)}] < J(\theta^{(t)})$ (\Cref{prop:fg-kendall}).

\item We prove that if the objective function is convex quadratic and the IGO algorithm instantiated with Gaussian distributions (i.e., the rank-$\mu$ update CMA-ES) is considered, $\E[J(\theta^{(t+1)}) \mid \theta^{(t)}] < J(\theta^{(t)})$ is satisfied by the IGO update using $g$ if the population Kendall's $\tau$ between $f$ and $g$ is sufficiently large (\Cref{thm:cq}).

\item We additionally investigate the use of the Pearson correlation coefficient between the weights assigned to candidate solutions based on $f$ and $g$ as an alternative approach to measure the quality of a surrogate function as an approximation of the objective function. It measures the difference between the parameter updates based on $f$ and $g$ more directly than the Kendall's rank correlation coefficient. The results derived are analogous to the ones derived for Kendall's rank correlation coefficient (\Cref{prop:fg-pearson,thm:cq:pearson}).
\end{itemize}

\section{Information-Geometric Optimization}\label{sec:igo}

As a representative example of a comparison-based search algorithm for the minimization of $f: \R^d \to \R$, we focus on the IGO framework \cite{igo}. 
Given a family $\mathcal{P}_\Theta$ of probability distributions parameterized by $\theta \in \Theta \subseteq \R^D$, the IGO framework provides the update rule for the distribution parameter $\theta$.
The population-based incremental learning (PBIL) \cite{pbil} and a variant of the CMA-ES, namely the rank-$\mu$ update CMA-ES, are derived from the IGO framework.
In this section, we describe the IGO framework and the IGO using the surrogate objective function $g: \R^d \to \R$.

\paragraph{IGO Update}

The iterative process of the IGO algorithm is described as follows: 
Suppose that $P_\theta$ admits the probability density $p(x;\theta)$ with respect to the Lebesgue measure on $\R^d$, and that the Fisher information matrix $F_{\theta} = \E[\nabla_\theta \ln p(x; \theta) \nabla_\theta \ln p(x; \theta)^\T]$ is non-singular for all $\theta \in \Theta$. 
Let $\theta^{(t)}$ be the parameter at iteration $t \geq 0$. 
At iteration $t$, the IGO samples $\lambda$ candidate solutions, $x_1, \dots,x_\lambda$, independently from $P_{\theta^{(t)}}$.
The candidate solutions are evaluated on objective function $f$. 
Let $f_i = f(x_i)$.
Then, their rankings are computed as
{\small\begin{align}
    r^{<}(f_i; \{f_k\}_{k=1}^{\lambda}) &= \sum_{j = 1}^{\lambda} \ind{f_j < f_i}, \label{eq:rlt}\\
    r^{\leq}(f_i; \{f_k\}_{k=1}^{\lambda}) &= \sum_{j = 1}^{\lambda} \ind{f_j \leq f_i}. \label{eq:rleq}
\end{align}}%
We define $r_i^< = r^{<}(f_i; \{f_k\}_{k=1}^{\lambda})$ and $r_i^{\leq} = r^{\leq}(f_i; \{f_k\}_{k=1}^{\lambda})$ for conciseness, where $r_i^<$ counts the number of strictly superior candidate solutions, and $r_i^{\leq}$ counts the number of better than or equally good candidate solutions. 
According to these rankings, the utility of each candidate solution is computed as
{\small\begin{align}
    W(f_i; \{f_k\}_{k=1}^{\lambda}) = \sum_{j=r_i^{<} + 1}^{r_i^{\leq}} \frac{w_j}{r_i^{\leq} - r_i^{<}} \enspace,\label{eq:w}
\end{align}}%
where $w_1, \dots, w_\lambda \in \R$ are the predefined weights.
For conciseness, we define $W_i = W(f_i; \{f_k\}_{k=1}^{\lambda})$.
If there is no tie, we have $r_{i}^{\leq} = r_{i}^{<} + 1$, and $W_i = w_{r_i^{\leq}}$.

The IGO update follows the weighted average of the natural gradients \cite{naturalgradient} of the log-likelihoods of the candidate solutions, namely $\theta^{(t+1)} = \theta^{(t)} + \alpha \cdot \Delta_f$, where $\alpha > 0$ is the learning rate and
{\small\begin{align}
    \Delta_f = \sum_{i=1}^{\lambda} W_i \cdot \tilde \nabla_\theta \ln p(x_i; \theta^{(t)}) .\label{eq:igo}
\end{align}}%
The natural gradient $\tilde \nabla_\theta \ln p(x; \theta)$ of the log-likelihood is the product of the inverse of the Fisher information matrix $F_{\theta}$ and the vanilla gradient $\nabla_\theta \ln p(x; \theta)$.%
\footnote{%
Here, we outline the intuition behind the IGO update \eqref{eq:igo}. 
The natural gradient of a function is known to be the steepest ascent direction of the function under the Fisher metric. In other words, it is the direction that the function value increases the most in the neighborhood where the KL-divergence from the current point is bounded by an infinitesimally small value.
Therefore, $\tilde \nabla_\theta \ln p(x_i; \theta^{(t)})$ is the direction in which the likelihood of $\theta$ at $x_i$ increases the most with respect to the KL-divergence. As \eqref{eq:igo} takes the weighted average of $\tilde \nabla_\theta \ln p(x_i; \theta^{(t)})$, where a greater weight is assigned to a candidate solution with a smaller objective value, it is clear that the IGO attempts to increase the likelihood of $\theta$ at candidate solutions with relatively smaller objective values.
}%

\paragraph{IGO Update with a Gaussian Distribution}

\Cref{sec:cq} focuses on the IGO algorithm instantiated using a family of Gaussian distributions with parameter $\theta = (m, \vect(C))$, where the probability density is
{\small\begin{equation}
    p(x; \theta) = \frac{1}{\sqrt{(2\pi)^d \det(C)}} \exp\left( - \frac{(x - m)^\T C^{-1} (x - m)}{2}\right) .
\end{equation}}%
Here, $\vect : \R^{d\times d} \to \R^{d^2}$ denotes the vectorization operator that converts a matrix into a column vector \cite{matrix}. 
The natural gradient of the log-likelihood can be written as $\tilde\nabla_\theta \ln p(x; \theta) = (\tilde\nabla_m \ln p(x; \theta), \tilde\nabla_C \ln p(x; \theta))$ \cite{nescma,cmang}, where
{\small\begin{align}
    \tilde\nabla_m \ln p(x; \theta) &= x - m,\label{eq:m}\\
    \tilde\nabla_C \ln p(x; \theta) &= \vect((x - m)(x - m)^\T - C).\label{eq:c}
\end{align}}%
This update corresponds to the rank-$\mu$ update CMA-ES.

\paragraph{IGO with Surrogate}

A surrogate objective $g:\R^d \to \R$ is often employed to save time owing to the computationally expensive evaluation of $f$.
In comparison-based approaches, the surrogate objective function is used to compute the rankings of candidate solutions. 
That is, $r^{<}(g_i; \{g_k\}_{k=1}^{\lambda})$ and $r^{\leq}(g_i; \{g_k\}_{k=1}^{\lambda})$ are computed instead of \eqref{eq:rlt} and \eqref{eq:rleq}, respectively.
The utility value $\tilde{W}_i = W(g_i; \{g_k\}_{k=1}^{\lambda})$ is then assigned to candidate solution $x_i$, where $r_i^<$ and $r_i^\leq$ in \eqref{eq:w} are replaced with $r^{<}(g_i; \{g_k\}_{k=1}^{\lambda})$ and $r^{\leq}(g_i; \{g_k\}_{k=1}^{\lambda})$, respectively.
The update of the IGO with the surrogate objective follows \eqref{eq:igo}, where $W_1,\dots,W_\lambda$ are replaced with $\tilde{W}_1,\dots,\tilde{W}_\lambda$, and $\theta^{(t+1)} = \theta^{(t)} + \alpha \cdot \Delta_g$, where
{\small\begin{align}
    \Delta_g = \sum_{i=1}^{\lambda} \tilde{W}_i \cdot \tilde \nabla_\theta \ln p(x_i; \theta^{(t)}) .\label{eq:igo-g}
\end{align}}

\section{Expected Objective Decrease}\label{sec:eod}

With or without a surrogate objective function $g$, the objective is to minimize the ground truth objective function $f$. 
That is, we want to locate the solution $x^* = \argmin_{x \in \R^d} f(x)$. 
If the family of probability distributions, $\mathcal{P}_\Theta$, includes a sequence of probability distributions that weakly converges to the Dirac-Delta distribution $\delta(x^*)$ concentrated at $x^*$, then
{\small\begin{equation}
    \inf_{\theta \in \Theta} \int f(x) p(x; \theta) \mathrm{d}x = f(x^*).
\end{equation}}
Therefore, by minimizing the expected objective function value $J(\theta) = \E[f(x)\mid\theta] = \int f(x) p(x; \theta) \mathrm{d}x$, we can achieve our goal. Hence, we use $\E_t[J(\theta^{(t+1)})] - J(\theta^{(t)})$ to measure the progress of the IGO update, where $\E_t$ denotes the expectation conditional on $\theta^{(t)}$.

In the following analysis, we assume:
\begin{assumption}\label{asm:1}
The expected objective $J(\theta)$ is continuously differentiable with respect to $\theta$, and there exists a non-negative definite symmetric matrix $H$ such that
{\small\begin{equation}
    J(\theta + \alpha\cdot \Delta) \leq J(\theta) + \alpha\cdot \nabla J(\theta)^\T \Delta + \frac{\alpha^2}{2}\Delta^\T H \Delta
\end{equation}}
for any $\Delta \in \R^D$, such that $\theta + \Delta \in \Theta$.
\end{assumption}
Under \Cref{asm:1}, letting $\Delta$ be either $\Delta_f$ or $\Delta_g$ and taking the expectation conditioned on $\theta^t$, we obtain the expected progress
{\small\begin{equation}
    \E_t[J(\theta^{(t+1)})] - J(\theta^{(t)}) \leq \alpha\nabla J(\theta^{(t)})^\T \E_t[\Delta] + \frac{\alpha^2}{2} \E_t[\Delta^\T H \Delta].\label{eq:drift}
\end{equation}}%

The second term on the right-hand side of \eqref{eq:drift} is bound from above by the following lemma. 
Without loss of generality, we consider $\Delta = \Delta_g$. 
For the case of $\Delta = \Delta_f$, we simply consider $g = f$. 
\begin{lemma}\label{lem:second}
Let $N_w = \lambda^2 \cdot \max_{k =1,\dots,\lambda}\{w_k^2\}$. 
For an arbitrary measurable $g$, 
{\small\begin{equation}
    \E[\Delta_g^\T H \Delta_g \mid \theta^{(t)} = \theta] \leq N_w \cdot \Tr(F_{\theta}^{-1}H).
\end{equation}}%
\end{lemma}
\begin{proof}
See \Cref{proof:lem:second}.
\end{proof}

From \Cref{lem:second} and \Cref{eq:drift}, we can immediately obtain a sufficient condition to guarantee the expected objective decrease in the IGO update with surrogate $g$.
\begin{proposition}\label{prop:diff}
Suppose that \Cref{asm:1} holds and that there exists a positive constant $\beta > 0$ such that $\nabla J(\theta^{(t)})^\T \E_t[\Delta_g] \leq - \beta \cdot N_w \cdot\Tr(F_{\theta^{(t)}}^{-1} H)$.
Then, for $\theta^{(t+1)} = \theta^{(t)} + \alpha \cdot \Delta_g$, 
{\small\begin{equation*}
    \E_t[J(\theta^{(t+1)})] - J(\theta^{(t)}) \leq \left(- \alpha\beta  + \frac{\alpha^2}{2} \right)\cdot N_w \cdot \Tr(F_{\theta^{(t)}}^{-1} H) .
\end{equation*}}%
\end{proposition}
If $\E_t[\Delta_g]$ is a descent direction of $J(\theta^{(t)})$, then there exists a sufficiently small learning rate $\alpha > 0$ such that $\E_t[J(\theta^{(t+1)})] < J(\theta^{(t)})$. 
However, $\alpha$ may depend on $\theta^{(t)}$, although \Cref{prop:diff} states that under the condition $\nabla J(\theta^{(t)})^\T \E_t[\Delta_g] \leq - \beta \cdot N_w \cdot \Tr(F_{\theta^{(t)}}^{-1} H)$ we can obtain $\E_t[J(\theta^{(t+1)})] < J(\theta^{(t)})$ with a learning rate $\alpha < 2 \beta$ that is independent of $\theta^{(t)}$. 

Though it is not trivial to prove that $\E_t[\Delta_f]$ is a descent direction of $J(\theta^{(t)})$ for a general $f$ because of the ranking-based weighting scheme, it is satisfied in some simple situations, as shown in \Cref{sec:cq}. Here, we assume that there exists a positive constant $\beta > 0$ such that 
{\small\begin{equation}
    \nabla J(\theta^{(t)})^\T \E_t[\Delta_f] \leq - \beta \cdot N_w \cdot \Tr(F_{\theta^{(t)}}^{-1} H).\label{eq:beta}
\end{equation}}%
Without assuming a sufficient improvement via a single step of the algorithm on the ground truth function $f$, we can not expect an improvement when using a surrogate function $g$. 
Thus, the main research question addressed in this paper is as follows: \emph{how ``close to $f$'' does surrogate $g$ need to be to guarantee $\nabla J(\theta^{(t)})^\T \E_t[\Delta_g] \leq - \gamma \cdot N_w \cdot \Tr(F_{\theta^{(t)}}^{-1} H)$ for some $\gamma > 0$?}

From \Cref{lem:second} and \Cref{eq:drift,eq:beta}, we have that the expected objective decrease in the IGO update with surrogate $g$ is bounded from above as
{\small\begin{equation}
\begin{split}
    &\E_t[J(\theta^{(t+1)})] - J(\theta^{(t)}) \\
    &\leq \alpha\nabla J(\theta^{(t)})^\T (\E_t[\Delta_f] + \E_t[\Delta_g-\Delta_f]) + \frac{\alpha^2}{2} \E_t[\Delta_g^\T H \Delta_g]\\
    &\leq \alpha\nabla J(\theta^{(t)})^\T\E_t[\Delta_g-\Delta_f] + \left(- \alpha\beta  + \frac{\alpha^2}{2} \right)\cdot N_w \cdot \Tr(F_{\theta^{(t)}}^{-1} H).
\end{split}\label{eq:gf-bound}
\end{equation}}%
To further bound the first term on the right-most side of the aforementioned inequality, 
we make the following assumption:
\begin{assumption}\label{asm:2}
Both $f$ and $g$ are measurable, and their level sets $\{f(x) = s\}$ and $\{g(x) = s\}$ have a zero Lebesgue measure for all $s \in \R$. 
\end{assumption}

The cumulative density of $g(x)$ being smaller than or equal to $s$ under $p(x; \theta)$ is defined as
 {\small\begin{equation*}
    P_g(s; \theta) = \int_{x: g(x) \leq s} p(x; \theta) \mathrm{d}x.
\end{equation*}}%
Proposition~1 of \cite{ode} shows that under \Cref{asm:2}, the expectation of $\tilde{W}_i$ given $x_i$ is
{\small\begin{equation*}
    \E_t[\tilde{W}_i \mid x_i] = \lambda^{-1} u(P_g(g(x_i); \theta^{(t)})),
\end{equation*}}%
where 
{\small\begin{equation*}
    u(p) = \lambda\sum_{i=1}^{\lambda} w_i \binom{\lambda - 1}{i - 1} p^{i-1}(1 - p)^{\lambda - i} .
\end{equation*}}%
Moreover, 
{\small\begin{align}
    \E_t[\Delta_g] = \E_{t}[ u(P_g(g(x); \theta^{(t)})) \cdot \tilde\nabla \ln p(x; \theta^{(t)}) ] . \label{eq:gtheta}
\end{align}}%
This holds for $f$ by simply replacing $g$ with $f$. 
Let 
{\small\begin{equation}
    K_w(\theta) = \E_{t}[(u(P_g(g(x); \theta)) - u(P_f(f(x); \theta)))^2]\label{eq:kw}
\end{equation}}%
be the expected squared difference between the utility values under $g$ and $f$. 
Then, we obtain the following result:
\begin{proposition}\label{prop:fg}
Suppose that Assumptions~\ref{asm:1} and \ref{asm:2} hold and that there exists a positive constant $\beta > 0$ such that \Cref{eq:beta} holds.
Let $K_w$ be defined in \eqref{eq:kw} and suppose that
{\small\begin{equation}
    \gamma = K_w(\theta^{(t)})^{1/2} \cdot \frac{ (\nabla J(\theta^{(t)})^\T F_{\theta^{(t)}}^{-1} \nabla J(\theta^{(t)}))^{1/2} }{ \abs{\nabla J(\theta^{(t)})^\T \E_t[\Delta_f] } } < 1.
\label{eq:gamma}
\end{equation}}%
Then, for $\theta^{(t+1)} = \theta^{(t)} + \alpha \cdot \Delta_g$, 
{\small\begin{equation*}
    \E_t[J(\theta^{(t+1)})] - J(\theta^{(t)}) \leq \left(- \alpha\beta(1-\gamma)  + \frac{\alpha^2}{2} \right)\cdot N_w \cdot \Tr(F_{\theta^{(t)}}^{-1} H) .
\end{equation*}}%
\end{proposition}
\begin{proof}
See \Cref{proof:prop:fg}. 
\end{proof}

The effect of the difference between the ground truth objective $f$ and its surrogate $g$ appears in $K_w(\theta)^{1/2}$, defined in \Cref{eq:kw}. If a surrogate $g$ is prepared such that
{\small\begin{equation*}
    K_w(\theta^{(t)})^{1/2} < \frac{\abs{\nabla J(\theta^{(t)})^\T \E_t[\Delta_f] }}{(\nabla J(\theta^{(t)})^\T F_{\theta^{(t)}}^{-1} \nabla J(\theta^{(t)}))^{1/2}},
\end{equation*}}%
then $\gamma < 1$, and hence, we can guarantee the expected objective decrease under $\alpha < 2 \beta (1 - \gamma)$.

\section{Kendall's Rank Correlation}\label{sec:kendall}

Kendall's rank correlation coefficient \cite{kendall} measures the similarity of the orderings of two sequences $\{f_i\}_{i=1}^{n}$ and $\{g_i\}_{i=1}^{n}$ and is computed as 
{\small\begin{equation}
    \hat{\tau} = \frac{n_c - n_d}{\sqrt{\binom{n}{2} - n_a} \sqrt{\binom{n}{2} - n_b}},
\end{equation}}%
where $n_a$, $n_b$, $n_c$, and $n_d$ are the numbers of pairs $(i,j)$ of indices among $\binom{n}{2}$ combinations such that
\begin{itemize}
\item[$n_a$:] $f_i = f_j$ (number of pairs with the same values in $\{f_i\}$);
\item[$n_b$:] $g_i = g_j$ (number of pairs with the same values in $\{g_i\}$);
\item[$n_c$:] $f_i < f_j$ and $g_i < g_j$, or $f_i > f_j$ and $g_i > g_j$ (concordant);
\item[$n_d$:] $f_i < f_j$ and $g_i > g_j$, or $f_i > f_j$ and $g_i < g_j$ (discordant).
\end{itemize}
If $\hat{\tau}$ is close to 1 or -1, the two sequences are strong-positively or strong-negatively correlated, respectively. If $\hat{\tau}$ is close to 0, they are considered to be uncorrelated. $\hat{\tau} = 1$ if and only if no pairs are discordant.

In this study, the two sequences correspond to the objective function values $\{f(x_i)\}$ and the surrogate function values $\{g(x_i)\}$ of candidate solutions $\{x_i\}$, generated independently from $P_{\theta}$. In such a case, $\tau$ is considered as an estimate of the population version of Kendall's rank correlation \cite{kendall}
{\small\begin{multline}
    \tau(\theta) = \Pr[(f(X) - f(\tilde{X}))(g(X) - g(\tilde{X})) > 0] \\
    - \Pr[(f(X) - f(\tilde{X}))(g(X) - g(\tilde{X})) < 0] ,\label{eq:pop-tau}
\end{multline}}%
where $X$ and $\tilde{X}$ are independently $P_\theta$-distributed.

Typically, the surrogate function $g$ is maintained to ensure that $\tau(\theta)$ is greater than or equal to a user-defined threshold $\bar{\tau} \in [-1, 1]$. 
Because $\tau(\theta)$ is not available in practice, it may be estimated using the moving average of $\hat{\tau}$. 
Hence, a second research question is as follows: \emph{can we guarantee the expected objective decrease if $g$ is maintained so that $\tau(\theta^{(t)}) \geq \bar{\tau}$?}

First, we show that we can bound $K_w(\theta)$, defined in \eqref{eq:kw}, using $\tau(\theta)$.
\begin{proposition}\label{prop:kendall}
Suppose \Cref{asm:2} holds.
Then, 
{\small\begin{equation*}
    \E_{x}[ ( u(P_f(f(x); \theta)) - u(P_g(g(x); \theta)) )^s]^{1/s} \leq L_u \cdot \left( \frac{1 - \tau(\theta))}{2}\right)^{1/s}
\end{equation*}}%
holds for any $s \geq 1$, where $L_u$ is the Lipschitz constant of $u$:
{\small\begin{equation*}
    L_u = \max_{0\leq p\leq 1}\abs*{\lambda(\lambda - 1) \sum_{k=1}^{\lambda-1}(- w_{k} + w_{k+1}) \binom{\lambda-2}{k-1} p^{k-1} (1 - p)^{\lambda - 1 - k}}.
\end{equation*}}%
\end{proposition}
\begin{proof} 
See \Cref{proof:prop:kendall}.
\end{proof}

\Cref{prop:kendall} with $s = 2$ implies that $K_w(\theta)^{1/2} \leq L_u (1/2 - \tau(\theta) / 2)^{1/2}$. 
Therefore, if we maintain $\tau(\theta^{(t)}) \geq \bar{\tau}$, then $K_w(\theta^{(t)})^{1/2} \leq L_u (1/2 - \bar{\tau} / 2)^{1/2}$. $\gamma$ in \eqref{eq:gamma} is bounded above by
{\small\begin{equation}
    \gamma_{\bar\tau} = L_u \left(\frac{1 - \bar{\tau}}{2} \right)^{1/2} \frac{(\nabla J(\theta^{(t)})^\T F_{\theta^{(t)}}^{-1} \nabla J(\theta^{(t)}))^{1/2}}{\abs{\nabla J(\theta^{(t)})^\T \E_t[\Delta_f] }} .\label{eq:gamma-tau}
\end{equation}}%
This is formally summarized as follows.
\begin{theorem}\label{prop:fg-kendall}
Suppose that Assumptions~\ref{asm:1} and \ref{asm:2} hold and that condition \eqref{eq:beta} is satisfied for all $\theta^{(t)} \in \Theta$.
Suppose that the surrogate $g$ is maintained so that $\tau(\theta^{(t)}) \geq \bar\tau$ and $\gamma_{\bar\tau} < 1$, where $\gamma_{\bar\tau}$ is defined in \eqref{eq:gamma-tau}.
Then, 
{\small\begin{equation}
    \E_t[J(\theta^{(t+1)})] - J(\theta^{(t)}) \leq \left(- \alpha\beta(1-\gamma_{\bar\tau})  + \frac{\alpha^2}{2} \right)\cdot N_w \Tr(F_{\theta^{(t)}}^{-1} H) .
\end{equation}}%
\end{theorem}
\begin{proof}
It is straightforward from \Cref{prop:fg,prop:kendall}.
\end{proof}

To guarantee a monotone decrease in the expected objective at any $\theta^{(t)} \in \Theta$, the threshold $\bar\tau$ needs to be set such that $\gamma_{\bar{\tau}} < 1$ for all $\theta^{(t)} \in \Theta$. 
That is, we need
{\small\begin{equation*}
\bar{\tau} > 1 - 2\inf_{\theta \in \Theta}\frac{\abs{\nabla J(\theta)^\T \E[\Delta_f\mid\theta] }^2}{L_u^2 (\nabla J(\theta)^\T F_{\theta}^{-1} \nabla J(\theta))} .
\end{equation*}}%
In general, it is difficult to evaluate the right-hand side of the aforementioned inequality. 
Moreover, the right-hand side of the inequality may equal 1, meaning that a monotone decrease in the expected objective can not be guaranteed unless the objective function $f$ and surrogate function $g$ are strictly concordant, i.e., $f(x) < f(y) \Leftrightarrow g(x) < g(y)$ for any $x, y \in \R^d$.

\section{Case Study: Convex Quadratic}\label{sec:cq}

We demonstrate that condition~\eqref{eq:beta} can be satisfied and that one can find $\bar\tau \in (-1, 1)$ such that $\gamma_{\bar\tau} < 1$. 
For this purpose, we focus on the IGO with the Gaussian model described in \Cref{eq:m,eq:c}. We consider a convex quadratic objective $f(x) = \frac{1}{2} (x - x^*)^\T A (x - x^*)$ and a surrogate function $g$ satisfying \Cref{asm:2}. 

First, we confirm that \Cref{asm:1} and \Cref{asm:2} hold. 
It is obvious that \Cref{asm:2} is satisfied with a convex quadratic function.
The expected objective function is
{\small\begin{equation}
    J(\theta) = \frac{1}{2}(m - x^*)^\T A (m - x^*) + \frac{1}{2} \Tr(A C) . 
    \label{eq:j}
\end{equation}}%
It is easy to see that \Cref{asm:1} is satisfied with block diagonal matrix $H = \diag(A, O)$, where $O$ is the $d^2 \times d^2$ dimensional zero matrix.

There are three terms we need to evaluate: $\nabla J(\theta^{(t)})^\T F_{\theta^{(t)}}^{-1} \nabla J(\theta^{(t)})$, $\Tr(F_{\theta^{(t)}}^{-1} H)$, and $\nabla J(\theta^{(t)})^\T \E_t[\Delta_f]$.
The gradient of $J$ is $\nabla_\theta J(\theta) = (\nabla_m J(\theta), \nabla_C J(\theta))$, where $ \nabla_m J(\theta) = A (m - x^*)$ and $\nabla_C J(\theta) = \frac12 \vect(A)$.
The inverse of the Fisher information matrix of $\theta$ is known from \cite{nescma} to be $F_{\theta}^{-1} = \diag(C, 2 C \otimes C)$.
Noting that $(C \otimes C) \vect(A) = \vect(CAC)$ and $\vect(A)^\T \vect(CAC) = \Tr(ACAC)$ \cite{matrix}, we have
{\small\begin{multline}
    \nabla J(\theta)^\T F_{\theta}^{-1} \nabla J(\theta) \\
        = (m - x^*) A C A (m - x^*) + \frac12 \Tr((AC)^2) =: M_f(\theta).\label{eq:mf}
\end{multline}}%
Moreover, we have $\Tr(F_{\theta}^{-1} H) = \Tr(C A) \leq (d \cdot \Tr((CA)^2))^{1/2} \leq (2d)^{1/2}M_f(\theta)^{1/2}$.

The evaluation of $\nabla J(\theta^{(t)})^\T\E_t[\Delta_f]$ is the most technical part; however, it is irrelevant to the use of surrogate function $g$. 
The main difficulty stems from the use of the ranking-based utility value. 
Using the techniques developed in \cite{ode}, we can bound it above as follows.
\begin{lemma}\label{lemma:f2}
Suppose that $w_i \geq w_{j}$ for all $i < j$ and $w_1 > w_{\lambda}$.
Let 
{\small\begin{align*}
    M_w := \sum_{k=1}^{\lambda} w_k \left(1 - \frac{2 k }{\lambda + 1}\right).
\end{align*}}%
Then, $M_w > 0$ and 
{\small\begin{equation*}
\nabla J(\theta^{(t)})^\T\E_t[\Delta_f] \leq - \frac{2^{1/2}}{6} M_w\cdot M_f(\theta^{(t)})^{1/2}. 
\end{equation*}}%
\end{lemma}
\begin{proof}
See \Cref{proof:lemma:f2}.
\end{proof}

Finally, we obtain the monotone decrease in the expected objective value under the IGO update with surrogate $g$. 
\begin{theorem}\label{thm:cq}
Suppose that $w_i \geq w_{j}$ for all $i < j$ and $w_1 > w_{\lambda}$.
Suppose that $f(x) = \frac{1}{2} (x - x^*)^\T A (x - x^*)$.
Let $\bar\tau \in \left(1 - \frac{M_w^2}{9L_u^2}, 1\right)$, where $M_w$ and $L_u$ are defined in \eqref{eq:mf} and \Cref{lemma:f2}, respectively. 
If all level sets of $g$ have a zero Lebesgue measure and $\tau(\theta^{(t)}) \geq \bar\tau$, then for any $\theta^{(t)} \in \Theta$, the update $\theta^{(t+1)} = \theta^{(t)} + \alpha \Delta_g$ leads to
{\small\begin{equation}
    \E_t[J(\theta^{(t+1)})] - J(\theta^{(t)}) \leq \left(- \alpha\beta(1-\gamma)  + \frac{\alpha^2}{2} \right)\cdot N_w \Tr(C A) ,\label{eq:cq:bound}
\end{equation}}%
where $\beta = \frac{M_w}{ 6 d^{1/2} N_w}$ and $\gamma = \frac{3 L_u(1 - \bar\tau)^{1/2}}{M_w}$.
\end{theorem}
\begin{proof}See \Cref{proof:thm:cq}.
\end{proof}

There are multiple implications of \Cref{thm:cq}. 
First, as long as the surrogate function $g$ is maintained to have the population version of Kendall's rank correlation between itself and $f$ greater than or equal to $\bar\tau$, IGO update $\theta^{(t+1)} = \theta^{(t)} + \alpha \Delta_g$ monotonically decreases the expected objective value. 
This fact justifies to some extent the use of Kendall's $\tau$ to decide whether the surrogate function $g$ should be used to rank candidate solutions or should be improved. 
Because a smooth function can be approximated locally around a local minimum using a convex quadratic function, the current analysis provides an insight into a local behavior of the IGO update with a surrogate on a smooth objective function.
Second, as already implied in \Cref{prop:fg-kendall}, a smaller $\bar{\tau}$ leads to a greater $\gamma$; hence, the learning rate needs to be smaller. 
To guarantee that the right-hand side of \eqref{eq:cq:bound} is negative, it is required that $\alpha < 2 \beta (1 - \gamma)$.
Moreover, the right-hand side of \eqref{eq:cq:bound} is minimized when $\alpha = \beta (1 - \gamma)$. 
This implies that the learning rate should be set smaller than the value used in the IGO update without a surrogate, and that it should be set smaller if $\bar\tau$ is set to be smaller. 
Using a smaller $\bar\tau$ and a smaller $\alpha$ leads to a smaller decrease in the expected objective value in terms of the number of iterations. 
However, this method may be useful if the computational cost of maintaining $g$ to keep $\tau(\theta) \geq \bar\tau$ is too high.

We will now discuss the related theoretical results in existing work. 
To the best of our knowledge, this is the first theoretical result showing the monotone decrease in the expected objective value under the rank-$\mu$ update CMA-ES using a surrogate function. 
For the case of the rank-$\mu$ update CMA-ES without a surrogate function, the monotone decrease in the expected objective value has been shown in \cite{cmang} for more general objective functions. On a convex quadratic function, the convergence rate and the convergence of the condition number of the product of the covariance matrix and the Hessian matrix to $1$ have been outlined in \cite{ngcq}. However, in \cite{cmang,ngcq}, the objective function value is used as the utility, and an infinite population size is assumed. 
That is, $\Delta_f = \nabla J(\theta^{(t)})$ with probability one, which significantly simplifies the analysis. 
The ranking-based utility is considered in \cite{igoml}, where the monotone decrease in the $q$-quantile of the objective function under a sampling distribution has been shown. 
However, in \cite{igoml}, an infinite population size is assumed. 
A finite population size is taken into account in \cite{igofinite} and the monotone decrease in the expected objective has been investigated. However, the objective function value is used as the utility and the isotropic Gaussian model, where $C$ is proportional to the identity matrix, is considered. A finite population size and the ranking-based utility are simultaneously considered in \cite{ode}, where the isotropic Gaussian model is used. The result for the IGO without surrogate in the present study is considered as an extension of the result for the general Gaussian model in \cite{ode}.

\section{Pearson Correlation on Weights}

The Pearson correlation coefficient is another popular metric for the similarity between two random variables $A$ and $B$.
We consider the Pearson correlation coefficient between $A = u(P_f(f(x); \theta))$ and $B = u(P_g(g(x); \theta))$ under $x \sim P_\theta$, given by
{\small\begin{equation}
    \rho(\theta) = \frac{\E[(A - \E[A])(B - \E[B])] }{\sqrt{\E[(A - \E[A])^2] } \cdot \sqrt{ \E[(B - \E[B])^2]}}.
\end{equation}}%
In practice, $\rho(\theta)$ may be approximated by the sample version of the Pearson correlation coefficient between $\{W_i\}_{i=1}^{n}$ and $\{\tilde{W}_i\}_{i=1}^{n}$, namely,
{\small\begin{equation}
    \hat{\rho} = \frac{\frac{1}{n}\sum_{i=1}^{n} (W_i - \langle W \rangle)(\tilde{W}_i - \langle \tilde{W} \rangle)}{\sqrt{\frac{1}{n}\sum_{i=1}^{n} (W_i - \langle W \rangle)^2} \sqrt{\frac{1}{n}\sum_{i=1}^{n} (\tilde{W}_i - \langle \tilde{W} \rangle)^2}},
\end{equation}}%
where $\langle W \rangle = \frac{1}{n}\sum_{i=1}^{n} W_i$ and $\langle \tilde{W} \rangle = \frac{1}{n}\sum_{i=1}^{n} \tilde{W}_i$.

Using the Pearson correlation coefficient, we can utilize the surrogate function more efficiently. 
If the weights are $w_1 = \dots = w_\mu = 1/\mu$ for $\mu \leq \lambda / 2$ and $w_{\mu+1} = \dots = w_{\lambda} = 0$, the parameter updates $\Delta_f$ and $\Delta_g$, computed using the objective function and the surrogate function, respectively, are the same as long as the better $\mu$ candidate solutions are correctly selected.
For example, consider the situation that the objective function values of $\lambda$ candidate solutions are $f_1 < \dots < f_\mu < f_{\mu + 1} < \dots < f_\lambda$ (i.e., $f_i < f_{i+1}$ for all $i=1,\dots,\lambda-1$), whereas the surrogate function values are $g_{\mu} < \dots < g_{1} < g_{\lambda} < \dots < g_{\mu+1}$ (i.e., $g_{i+1} < g_{i}$ for $i=1,\dots,\mu-1$, $g_{1} < g_{\mu+1}$, and $g_{j} < g_{j+1}$ for $j = \mu+1,\dots,\lambda$). 
As mentioned previously, we have $\Delta_f = \Delta_g$; hence, $\Delta_g$ can be used with no risk. 
In this case, the Pearson correlation coefficient is $\rho(\theta) = 1$. Therefore, $\rho(\theta) \geq \bar\rho$ is satisfied with any $\bar\rho < 1$, and $\Delta_g$ will be used. Conversely, Kendall's rank correlation coefficient is $\tau(\theta) = - [(\lambda - 2 \mu)^2 + \lambda] / (\lambda^2 + \lambda) < 0$. One cannot use $\Delta_g$ unless $\bar\tau$ is set to a negative value, which is unreasonable.

The Pearson correlation coefficient is more directly related to $K_w$ in \Cref{eq:kw} than Kendall's rank correlation coefficient. The same $\rho(\theta)$ value results in the same $K_w(\theta)$, whereas the same $\tau(\theta)$ values do not necessarily result in the same $K_w(\theta)$. 
\begin{proposition}\label{prop:pearson}
Suppose \Cref{asm:2} holds.
Let 
{\small\begin{equation*}
    U_u = \sum_{i=1}^{\lambda} \sum_{i=1}^{\lambda}w_i w_j \left( \frac{\lambda^2}{2\lambda - 1}\binom{\lambda-1}{i-1} \binom{\lambda-1}{j-1} \binom{2\lambda-2}{i+j-2}^{-1} - 1\right).
\end{equation*}}%
Then, $K_w(\theta) = 2 \cdot U_u \cdot (1 - \rho(\theta))$.
\end{proposition}
\begin{proof} See \Cref{proof:prop:pearson}.
\end{proof}

We can obtain the counterparts of \Cref{prop:fg-kendall,thm:cq} as follows.
\begin{theorem}\label{prop:fg-pearson}
Suppose that Assumptions~\ref{asm:1} and \ref{asm:2} hold and that condition \eqref{eq:beta} is satisfied.
If the surrogate $g$ is maintained so that $\rho(\theta^{(t)}) \geq \bar\rho$, we have 
{\small\begin{equation}
    \E_t[J(\theta^{(t+1)})] - J(\theta^{(t)}) \leq \left(- \alpha\beta(1-\gamma_{\bar\rho})  + \frac{\alpha^2}{2} \right)\cdot N_w \Tr(F_{\theta^{(t)}}^{-1} H) ,
\end{equation}}%
where $\gamma_{\bar\rho}$ is defined as
{\small\begin{equation}
    \gamma_{\bar\rho} = (2 U_u\cdot (1 - \bar{\rho}) )^{1/2} \frac{(\nabla J(\theta^{(t)})^\T F_{\theta^{(t)}}^{-1} \nabla J(\theta^{(t)}))^{1/2}}{\abs{\nabla J(\theta^{(t)})^\T \E_t[\Delta_f] }} .\label{eq:gamma-rho}
\end{equation}}%

\end{theorem}
\begin{proof}
It is straightforward from \Cref{prop:fg,prop:pearson}.
\end{proof}

\begin{theorem}\label{thm:cq:pearson}
Suppose that $w_i \geq w_{j}$ for all $i < j$ and $w_1 > w_{\lambda}$.
Suppose that $f(x) = \frac{1}{2} (x - x^*)^\T A (x - x^*)$.
Let $\bar\rho \in \left(1 - \frac{M_w^2}{36 U_u}, 1\right)$, where $M_w$ and $U_u$ are defined in \eqref{eq:mf} and \Cref{prop:pearson}, respectively. 
If all level sets of $g$ have a zero Lebesgue measure and $\rho(\theta^{(t)}) \geq \bar\rho$, then for any $\theta^{(t)} \in \Theta$, the update $\theta^{(t+1)} = \theta^{(t)} + \alpha \Delta_g$ leads to \Cref{eq:cq:bound} with
$\beta = \frac{M_w}{ 6 d^{1/2} N_w}$ and $\gamma = \frac{6 U_u^{1/2} (1 - \bar{\rho})^{1/2}}{M_w}$.
\end{theorem}
\begin{proof}The proof is analogous to the proof of \Cref{thm:cq}.
\end{proof}


\section{Conclusion}

In this paper, we present an investigation of the IGO algorithm using a surrogate function.
First, we concluded that if condition~\eqref{eq:beta} and condition~\eqref{eq:gamma} are satisfied for some $\beta > 0$ and $\gamma \in (0, 1)$, respectively, the IGO update $\theta^{(t+1)} = \theta^{(t)} + \alpha \Delta_g$ using a surrogate function decreases the expected objective function value, i.e., $\E[J(\theta^{(t+1)})] < \E[J(\theta^{(t)})]$ (\Cref{prop:fg}).
Second, we showed that the condition~\eqref{eq:gamma} is satisfied if the population version of Kendall's rank correlation coefficient between the objective function value and the surrogate function value is sufficiently close to one (\Cref{prop:fg-kendall}). 
Third, by considering the IGO algorithm instantiated by the Gaussian distributions, i.e., the rank-$\mu$ update CMA-ES, we have shown that both condition~\eqref{eq:beta} and condition~\eqref{eq:gamma} are satisfied on a convex quadratic objective function if the surrogate function is maintained to have the population version of Kendall's rank correlation coefficient greater than a constant value. 
Fourth, we investigated the use of the Pearson correlation coefficient between the weights $\{W_i\}$ assigned to candidate solutions based on the ground truth objective $f$ and the weights $\{\tilde{W}_i\}$ based on the surrogate $g$, as it can more directly control the difference between the update of the distribution parameters.

Our results partly justify the current use of surrogate functions. 
Practical surrogate-assisted approaches, e.g., \cite{lqcmaes}, maintains a surrogate function so that the Kendall's rank correlation coefficient is estimated to be greater than a predefined threshold. 
If the threshold is sufficiently close to one, this approach will lead to a monotonic decrease in the expected objective function value. 

We end this paper by stating the limitations and describing possible future works. First, the optimality of our theoretical results is to be investigated. We conjecture that the threshold values $\bar\tau$ and $\bar\rho$ to guarantee a monotonic decrease in the expected objective function on a convex quadratic problem are not optimal. 
This is partly because it is difficult to obtain an optimal bound in \eqref{eq:beta} even without a surrogate function because of the ranking-based weighting scheme. 
For practical uses, we would like to set the threshold sufficiently small to exploit a surrogate function as long as we can derive a monotonic improvement. 
The optimality must be investigated both empirically and theoretically. 
Second, the current analysis does not reveal how many objective function evaluations can be saved by utilizing a surrogate function. 
To analyze this, one needs to estimate the required number of objective function evaluations to maintain a surrogate function. 
Third, the current analysis does not suggest how a surrogate function should be trained or maintained so that the Kendall's rank correlation coefficient between the surrogate function values and the ground truth objective function values is sufficiently high. 
In practice, it is impossible to check whether the criterion is satisfied under the limited knowledge of the objective function. Therefore, we need to approximate the criterion by requiring several $f$-calls. The second and third points are very important in practical situations and must be theoretically investigated in future studies.
Finally, the proposed use of Pearson's correlation coefficient is to be investigated empirically.


\begin{acks}
This research is partially supported by the JSPS KAKENHI Grant Number 19H04179.
\end{acks}

\appendix

\section{Proof}

\subsection{Proof of \Cref{lem:second}}\label{proof:lem:second}

Let $\norm{\Delta}_H^2 = \Delta^\T H \Delta$. 
Noting that $\abs{\tilde{W}_i} \leq \max_{k =1,\dots,\lambda} \{\abs{w_i}\}$, we obtain
{\small\begin{align*}
    \E_t\left[ \Delta_g^\T H \Delta_g \right]
    &= \E_t\left[ \norm*{\sum_{i=1}^{\lambda} \tilde{W}_i \cdot \tilde \nabla_\theta \ln p(x_i; \theta^{(t)})}_H^2 \right] \\
    &\leq \E_t\left[ \left(\sum_{i=1}^{\lambda} \abs{\tilde{W}_i} \cdot \norm{\tilde \nabla_\theta \ln p(x_i; \theta^{(t)})}_H \right)^2 \right] \\
    &\leq \max_{k =1,\dots,\lambda} \{w_i^2\} \E_t\left[ \left(\sum_{i=1}^{\lambda} \norm{\tilde \nabla_\theta \ln p(x_i; \theta^{(t)})}_H \right)^2 \right] \\
    &\leq \lambda^2 \max_{k =1,\dots,\lambda} \{w_i^2\} \E_t\left[ \norm{\tilde \nabla_\theta \ln p(x_i; \theta^{(t)})}_H^2 \right] \\
    &= \lambda^2 \max_{k =1,\dots,\lambda} \{w_i^2\}  \Tr(F_{\theta^{(t)}}^{-1}H) .
\end{align*}}%
The last equality used the following:
{\small\begin{align*}
    \MoveEqLeft[2]
    \E_t\left[ \norm{\tilde \nabla_\theta \ln p(x_i; \theta^{(t)})}_H^2 \right]\\
    &= 
    \E_t\left[ \nabla_\theta \ln p(x_i; \theta^{(t)})^{\T} F_{\theta^{(t)}}^{-1} H F_{\theta^{(t)}}^{-1} \nabla_\theta \ln p(x_i; \theta^{(t)}) \right]\\
    &= 
    \E_t\left[ \Tr\left(\nabla_\theta \ln p(x_i; \theta^{(t)})^{\T} F_{\theta^{(t)}}^{-1} H F_{\theta^{(t)}}^{-1} \nabla_\theta \ln p(x_i; \theta^{(t)})\right) \right]\\
    &= 
    \E_t\left[ \Tr\left(F_{\theta^{(t)}}^{-1} H F_{\theta^{(t)}}^{-1} \nabla_\theta \ln p(x_i; \theta^{(t)}) \nabla_\theta \ln p(x_i; \theta^{(t)})^{\T} \right) \right]\\    
    &= 
    \Tr\left(F_{\theta^{(t)}}^{-1} H F_{\theta^{(t)}}^{-1} \E_t\left[ \nabla_\theta \ln p(x_i; \theta^{(t)}) \nabla_\theta \ln p(x_i; \theta^{(t)})^{\T} \right]\right)\\
    &= 
    \Tr\left(F_{\theta^{(t)}}^{-1} H F_{\theta^{(t)}}^{-1} F_{\theta^{(t)}}\right) 
    \\
    &= 
    \Tr\left(F_{\theta^{(t)}}^{-1} H \right) .
\end{align*}}%
This completes the proof.\qed

\subsection{Proof of \Cref{prop:fg}}\label{proof:prop:fg}

In light of \Cref{eq:gf-bound}, it is sufficient to show that
{\small\begin{equation*}
    \nabla J(\theta^{(t)})^\T\E_t[\Delta_g-\Delta_f]
    \leq \beta \cdot \gamma \cdot N_w \cdot \Tr(F_{\theta^{(t)}}^{-1} H).
\end{equation*}}%
Because of conditions \eqref{eq:beta} and \eqref{eq:gamma}, it suffices to show that $\nabla J(\theta^{(t)})^\T\E_t[\Delta_g-\Delta_f] \leq K_w(\theta^{(t)})^{1/2} \cdot (\nabla J(\theta^{(t)})^\T F_{\theta^{(t)}}^{-1} \nabla J(\theta^{(t)}))^{1/2}$.

In light of \Cref{eq:gtheta,eq:kw}, we have
{\small\begin{align*}
    \MoveEqLeft[2]\nabla J(\theta^{(t)})^\T\E_t[\Delta_g-\Delta_f]\\
    &= \begin{aligned}[t]
    \nabla J(\theta^{(t)})^\T\E_{t}[&(u(P_g(g(x); \theta^{(t)})) - u(P_f(f(x); \theta^{(t)}))) \\
    &\cdot \tilde\nabla \ln p(x; \theta^{(t)}) ]
    \end{aligned}\\
    &= \begin{aligned}[t]
    \E_{t}[&(u(P_g(g(x); \theta^{(t)})) - u(P_f(f(x); \theta^{(t)}))) \\
    &\cdot \nabla J(\theta^{(t)})^\T \tilde\nabla \ln p(x; \theta^{(t)}) ]
    \end{aligned}\\
    &\leq \begin{aligned}[t]
    &\E_{t}[(u(P_g(g(x); \theta^{(t)})) - u(P_f(f(x); \theta^{(t)})))^2]^{1/2} \\
    &\cdot \E_{t}[ ( \nabla J(\theta^{(t)})^\T \tilde\nabla \ln p(x; \theta^{(t)}))^2 ]^{1/2} 
    \end{aligned}\\
    &= K_w(\theta^{(t)})^{1/2} \cdot (\nabla J(\theta^{(t)})^\T F_{\theta^{(t)}}^{-1} \nabla J(\theta^{(t)}))^{1/2},
\end{align*}}%
where we used
{\small\begin{align*}
    \MoveEqLeft[2]\E_{t}[ ( \nabla J(\theta^{(t)})^\T \tilde\nabla \ln p(x; \theta^{(t)}))^2 ]\\
    &= \E_{t}[ \nabla J(\theta^{(t)})^\T \tilde\nabla \ln p(x; \theta^{(t)})\tilde\nabla \ln p(x; \theta^{(t)})^\T \nabla J(\theta^{(t)}) ] \\
    &= \nabla J(\theta^{(t)})^\T \E_t[\tilde\nabla \ln p(x; \theta^{(t)})\tilde\nabla \ln p(x; \theta^{(t)})^\T] \nabla J(\theta^{(t)}) \\
    &= \nabla J(\theta^{(t)})^\T F_{\theta^{(t)}}^{-1} F_{\theta^{(t)}} F_{\theta^{(t)}}^{-1} \nabla J(\theta^{(t)}) 
    = \nabla J(\theta^{(t)})^\T F_{\theta^{(t)}}^{-1} \nabla J(\theta^{(t)}) .
\end{align*}}%
This completes the proof.\qed

\subsection{Proof of \Cref{prop:kendall}}\label{proof:prop:kendall}

Note that the derivative of $u$ is 
{\small\begin{equation}
    \frac{\mathrm{d}u(p)}{\mathrm{d}p} = \lambda(\lambda - 1) \sum_{k=1}^{\lambda-1}(- w_{k} + w_{k+1}) \binom{\lambda-2}{k-1} p^{k-1} (1 - p)^{\lambda - 1 - k}.\label{eq:du}
\end{equation}}%
Therefore, $u$ is $L_u$-Lipschitz continuous. (Its trivial upper bound is $L_u \leq \lambda(\lambda - 1)\max_{k \in \llbracket 1, \lambda-1\rrbracket}\abs{w_{k+1}-w_{k}}$ because $\sum_{k=1}^{\lambda-1}\binom{\lambda-2}{k-1} p^{k-1} (1 - p)^{\lambda - 1 - k} = 1$). 
Under \Cref{asm:2}, the second term of \eqref{eq:pop-tau} is one minus the first term; hence,
{\small\begin{align*}
    \tau(\theta) 
    =& 2 \Pr[(f(X) - f(\tilde{X}))(g(X) - g(\tilde{X})) > 0] - 1 \\
    =& 2 \E[\ind{f(X) > f(\tilde{X})}\ind{g(X) > g(\tilde{X})} \\
    &+ \ind{f(X) < f(\tilde{X})}\ind{g(X) < g(\tilde{X})}] - 1 \\
    =& 2 \E[\ind{f(X) > f(\tilde{X})}\ind{g(X) > g(\tilde{X})} \\
    &+ (1-\ind{f(X) > f(\tilde{X})})(1-\ind{g(X) > g(\tilde{X})})] - 1 \\
    =& 4 \E[\ind{f(X) > f(\tilde{X})}\ind{g(X) > g(\tilde{X})}] + 2 \\
    &- 2 \E[\ind{f(X) > f(\tilde{X})}] - 2\E[\ind{g(X) > g(\tilde{X})}] - 1 \\
    =& 4 \E[\ind{f(X) > f(\tilde{X})}\ind{g(X) > g(\tilde{X})}] - 1.
\end{align*}}%
Then, for any $s \geq 1$,
{\small\begin{align*}
\MoveEqLeft[1]\E_{x}[ ( u(P_f(f(x); \theta)) - u(P_g(g(x); \theta)) )^s]^{1/s}\\
    =& \E_{x}[ ( u(\E[\ind{f(X) < f(x)}]) - u(\E[\ind{g(X) < g(x)}]) )^s]^{1/s} \\
    \leq& L_u \E_{x}[ ( \E[\ind{f(X) < f(x)}] - \E[\ind{g(X) < g(x)}] )^s]^{1/s} \\
    \leq& L_u \E_{x}[ \abs{ \E[\ind{f(X) < f(x)}] - \E[\ind{g(X) < g(x)}] } ]^{1/s} \\
    =& L_u \E_{x}[ \abs{ \E[\ind{f(X) < f(x)} - \ind{g(X) < g(x)}] } ]^{1/s} \\
    \leq& L_u \E_{x}[ \E[ \abs{ \ind{f(X) < f(x)} - \ind{g(X) < g(x)} } ]]^{1/s} \\
    =& L_u \E_{x}[ \E[ ( \ind{f(X) < f(x)} - \ind{g(X) < g(x)} )^2 ]]^{1/s} \\
    =& L_u \E_{x}[ \E[ \ind{f(X) < f(x)} + \ind{g(X) < g(x)} \\
    &- 2 \ind{f(X) < f(x)} \ind{g(X) < g(x)} ]]^{1/s} \\
    =& L_u (1 - 2 \E_{x}[ \E[ \ind{f(X) < f(x)} \ind{g(X) < g(x)} ]])^{1/s} \\
    =& L_u (1/2 - \tau(\theta) / 2)^{1/s}. 
\end{align*}}%
This completes the proof.\qed

\subsection{Proof of \Cref{lemma:f2}}\label{proof:lemma:f2}
First, note that
{\small\begin{align*}
\MoveEqLeft[2]\nabla J(\theta^{(t)})^\T \tilde\nabla \ln p(x; \theta^{(t)})\\
&= (m - x^*)^\T A (x - m) + \frac12 \Tr(A ((x - m)(x - m)^\T - C)) \\
&= f(x) - \E_t[f(x)].
\end{align*}}%
Using this equality, we obtain
{\small\begin{align*}
\MoveEqLeft[2]\nabla J(\theta^{(t)})^\T\E_t[\Delta_f] \\
&= \nabla J(\theta^{(t)})^\T \E_{t}[ u(P_f(f(x); \theta^{(t)})) \cdot \tilde\nabla \ln p(x; \theta^{(t)}) ] \\
&=  \E_{t}[ u(P_f(f(x); \theta^{(t)})) \cdot \nabla J(\theta^{(t)})^\T \tilde\nabla \ln p(x; \theta^{(t)}) ]\\
&=  \E_{t}[ u(P_f(f(x); \theta^{(t)})) \cdot (f(x) - \E_t[f(x)]) ].
\end{align*}}%

We set an upper bound on the right-most side of the aforementioned equality.
Under the assumptions on $w_1,\dots,w_\lambda$, $\frac{\mathrm{d}u(p)}{\mathrm{d}p}$ is negative because each term in \eqref{eq:du} is non-positive due to the condition $w_i \geq w_{i+1}$, and at least one term is negative due to $w_1 > w_\lambda$. That is, $u$ is strictly decreasing in $[0, 1]$. 
Then, we see that $u(P_f(f(x); \theta^{(t)}))$ and $(f(x) - \E_t[f(x)])$ are negatively correlated, and the improved Chebyshev sum inequality (Theorem~21 of \cite{ode}) reads
{\small\begin{multline*}
\E_t[u(P_f(f(x); \theta^{(t)})) \cdot (f(x) - \E_t[f(x)])] 
\\ \leq - \frac{1}{4} \E_t[ \abs{u(P_f(f(x); \theta^{(t)})) - u(P_f(f(y); \theta^{(t)}))}] \cdot \E_t[\abs{f(x) - f(y)}] 
\end{multline*}}%
for $x$ and $y$ being independently $P_{\theta^{(t)}}$-distributed.
Proposition~23 of \cite{ode} shows that
{\small\begin{equation}
\E_t[ \abs{u(P_f(f(x); \theta^{(t)})) - u(P_f(f(y); \theta^{(t)}))}]
= 2 M_w.\label{eq:mw}
\end{equation}}%
To bound $\E[\abs{f(x) - f(y)}]$ below, we use the so-called fourth-moment method (Proposition~22 of \cite{ode}) and obtain
{\small\begin{multline}
\E[\abs{f(x) - f(y)}]  \\
\geq 2 \left( \frac{\E[(f(x) - \E[f(x)])^2]^3}{\E[(f(x) - \E[f(x)])^4] + 3 \E[(f(x) - \E[f(x)])^2]^2} \right)^{1/2}.\label{eq:4th-moment}
\end{multline}}%
Let the eigenvalue decomposition of $\sqrt{C} A \sqrt{C}$ be denoted by $E D E^\T$, where $D$ is the diagonal matrix composed of the square root of the eigenvalues of $\sqrt{C} A \sqrt{C}$, and $E$ is the orthogonal matrix composed of the unit eigenvectors of $\sqrt{C} A \sqrt{C}$. Let $z = E^\T \sqrt{C}^{-1}(x - m)$ and $v = E^\T \sqrt{C} A (m - x^*)$. Then, we have
{\small\begin{align*}
    &f(x) - E[f(x)]\\
    &= \frac12 (x - m)^\T A (x - m) - \frac12 \Tr(A C) + (m - x^*)^\T A (x - m) \\
    &= \frac12 z^\T E^\T \sqrt{C} A \sqrt{C} E z - \frac12 \Tr( \sqrt{C} A \sqrt{C}) + v^\T E^\T \sqrt{C}^{-1} A^{-1} A \sqrt{C} E z \\
    &= \frac12 z^\T D z - \frac12 \Tr( D ) + v^\T z 
    \\
    &= \frac12 \sum_{i=1}^{d} \left( d_i (z_i^2 - 1) + 2 v_i z_i \right).
\end{align*}}%
Let
{\small\begin{align*}
    \mu_{i,2} = \E_t\left[ \left( d_i (z_i^2 - 1) + 2 v_i z_i \right)^2 \right] &= 2 d_i^2 + 4 v_i^2, \\
    \mu_{i,4} = \E_t\left[ \left( d_i (z_i^2 - 1) + 2 v_i z_i \right)^4 \right] &= 60 d_i^4 + 240 d_i^2 v_i^2 + 48 v_i^4.
\end{align*}}%
Note that $z_i$ are independently and standard normally distributed. A simple derivation leads to
{\small\begin{align*}
    \E[(f(x) - \E[f(x)])^2] = \frac14 \sum_{i=1}^{d} \mu_{i,2}
\end{align*}}%
and
{\small\begin{equation*}
    \E[(f(x) - \E[f(x)])^4]
    =     \frac{1}{16}\sum_{i=1}^{d}\left( \mu_{i,4} 
    - 3 \mu_{i,2}^2 \right) + \frac{3}{16}\left( \sum_{i=1}^d \mu_{i,2} \right)^2 .
\end{equation*}}%
Then,
{\small\begin{equation}
\begin{split}
\MoveEqLeft[2]\frac{\E[(f(x) - \E[f(x)])^2]^3}{\E[(f(x) - \E[f(x)])^4] + 3 \E[(f(x) - \E[f(x)])^2]^2} \\
&=
\frac{ \left( \sum_{i=1}^{d} \mu_{i,2} \right) }{ 4 \frac{\sum_{i=1}^{d}\left( \mu_{i,4} 
    - 3 \mu_{i,2}^2 \right)}{ \left( \sum_{i=1}^d \mu_{i,2} \right)^2 } + 24}
    \\
&\geq \frac{1}{72} \left( \sum_{i=1}^{d} \mu_{i,2} \right) \\
&= \frac{1}{36} \left( \sum_{i=1}^{d} d_i^2 + 2 v_i^2 \right), \label{eq:mf-bound}
\end{split}
\end{equation}}%
where we used $\mu_{i,4} - 3 \mu_{i,2}^2 \leq 12 \mu_{i,2}^2$. 
We note that
{\small\begin{equation}
    \sum_{i=1}^{d} d_i^2+ 2 v_i^2 = 2 M_f. \label{eq:mf-relation}
\end{equation}}%
From \eqref{eq:mw}, \eqref{eq:4th-moment}, \eqref{eq:mf-bound} and \eqref{eq:mf-relation}, the desired inequality is obtained. \qed

\subsection{Proof of \Cref{thm:cq}}\label{proof:thm:cq}
In light of \Cref{lemma:f2}, we have 
{\small\begin{align*}
 \nabla J(\theta^{(t)})^\T \E_t[\Delta_f]
 &\leq - \frac{2^{1/2}}{6} M_w\cdot M_f^{1/2}\\
 &\leq -\frac{1}{6 d^{1/2}} M_w \cdot \Tr(F_{\theta^{(t)}}^{-1} H),
\end{align*}}%
where we used $\Tr(F_{\theta}^{-1} H) \leq (2d)^{1/2}M_f^{1/2}$. Therefore, condition~\eqref{eq:beta} is satisfied with $\beta = \frac{M_w}{ 6 d^{1/2} N_w}$. 

Again, in light of \Cref{lemma:f2}, we have
{\small\begin{align*}
 \abs{\nabla J(\theta^{(t)})^\T \E_t[\Delta_f]}
 &\geq \frac{2^{1/2}}{6} M_w\cdot M_f^{1/2}
 \\
 &= \frac{2^{1/2}}{6} M_w\cdot (\nabla J(\theta^{(t)})^\T F_{\theta^{(t)}}^{-1} \nabla J(\theta^{(t)}))^{1/2},
\end{align*}}%
where we used $M_f = \nabla J(\theta^{(t)})^\T F_{\theta^{(t)}}^{-1} \nabla J(\theta^{(t)})$. 
Then, the right-hand side of \eqref{eq:gamma-tau} is bounded above by
{\small\begin{equation}
 \frac{3 L_u (1 - \bar{\tau})^{1/2}}{M_w}. \end{equation}}%
Therefore, condition~\eqref{eq:gamma} is satisfied with $\gamma =  \frac{3 L_u (1 - \bar{\tau})^{1/2}}{M_w}$.
This completes the proof.
\qed

\subsection{Proof of \Cref{prop:pearson}}\label{proof:prop:pearson}
First, using the formula $\int_0^1 p^{n}(1-p)^{n-k} \mathrm{d}p = (n+1)^{-1} \binom{n}{k}^{-1}$, we have $\int_0^1 u(t) \mathrm{d}t = \sum_{i=1}^{\lambda}w_i$ and
{\small\begin{gather*}
    \int_{0}^{1}u^2(t) \mathrm{d}t = 
    \frac{\lambda^2}{2\lambda - 1}
    \sum_{i=1}^{\lambda} \sum_{i=1}^{\lambda}w_i w_j \binom{\lambda-1}{i-1} \binom{\lambda-1}{j-1} \binom{2\lambda-2}{i+j-2}^{-1}.
\end{gather*}}%
Therefore, $U_u = \int_{0}^{1}u^2(t) \mathrm{d}t - \left(\int_{0}^{1}u(t) \mathrm{d}t \right)^2$. 

Under \Cref{asm:2}, both $f(x)$ and $g(x)$ have continuous cumulative density functions under $x \sim P_{\theta}$. In light of \cite[Theorem~2.1]{nonuniform}, both $P_f(f(x); \theta)$ and $P_g(g(x); \theta))$ are uniformly distributed on $[0, 1]$. Therefore, we have
{\small\begin{gather*}
    \E[A] = \E[B] = \int_{0}^{1} u(t) \mathrm{d}t, \\
    \E[A^2] = \E[B^2] = \int_{0}^{1} u^2(t) \mathrm{d}t.
\end{gather*}}%
Hence, we have
{\small\begin{equation*}
    \E\left[\left(A - \sum_{i=1}^{\lambda}w_i\right)\left(B - \sum_{i=1}^{\lambda}w_i\right)\right] = U_u \cdot \rho(\theta) . 
\end{equation*}}%

We finally obtain
{\small\begin{align*}
    \MoveEqLeft[0]\E_{x}[ ( u(P_f(f(x); \theta)) - u(P_g(g(x); \theta)) )^2]\\
    &=\E[ ( A - B )^2]
    =\E[ ((A - \E[A]) - (B - \E[B]) )^2 ]\\
    &=\E[ (A - \E[A])^2 ] + \E[ (B - \E[B])^2 ] 
     - 2 \E[ (A - \E[A])(B - \E[B]) ]\\
    &= 2 U_u - 2 U_2 \rho(\theta) 
    = 2 U_u (1 - \rho(\theta)).  
\end{align*}}%
This completes the proof.\qed



\end{document}